\documentclass[journal,twoside,web]{ieeecolor}
\usepackage{graphicx}
\usepackage{epstopdf}
\usepackage{tmi}
\usepackage{cite}
\usepackage{amsmath,amssymb,amsfonts}
\usepackage{algorithmic}
\usepackage{graphicx}
\usepackage{textcomp}
\usepackage{booktabs}  
\usepackage{microtype}   
\usepackage[svgnames]{xcolor} 
\usepackage{diagbox}
\usepackage{multirow}
\usepackage{array, makecell}
\usepackage{subcaption}

\definecolor{Teal}{RGB}{0,0.541,0.855}

\usepackage{hyperref}       
\hypersetup{
    colorlinks=true,
    citecolor=blue,
    linkcolor=subsectioncolor,
    filecolor=magenta,      
    urlcolor=cyan,
    }
\usepackage[nameinlink,capitalize]{cleveref}

\DeclareMathOperator*{\argmin}{arg\,min}

\Crefformat{equation}{#2Eq.~(#1)#3}
\Crefformat{figure}{#2Fig.~#1#3}
\Crefmultiformat{figure}{Figs.~#2#1#3}{ and~#2#1#3}{,~#2#1#3}{ and~#2#1#3}

\def\BibTeX{{\rm B\kern-.05em{\sc i\kern-.025em b}\kern-.08em
    T\kern-.1667em\lower.7ex\hbox{E}\kern-.125emX}}
\begin{document}
\title{SC-MIL: Sparsely Coding Multiple Instance Learning for Whole Slide Image Classification}
\author{Peijie Qiu, Pan Xiao, Wenhui Zhu, Yalin Wang, and  Aristeidis Sotiras
\thanks{P. Qiu and P. Xiao are with Mallinckrodt Institute of Radiology, Washington University School of Medicine, St. Louis, MO 63110 USA (e-mail: \{peijie.qiu, panxiao\}@wustl.edu). }
\thanks{W. Zhu and Y. Wang are with School of Computing and Augmented Intelligence,
Arizona State University, Tempe, AZ 85281 USA (e-mail: \{wzhu59, ylwang\}@asu.edu). }
\thanks{A. Sotiras is with Mallinckrodt Institute of Radiology and the Institute for Informatics, Data Science and Biostatistics, 
Washington University School of Medicine, St. Louis, MO 63110 USA (e-mail: aristeidis.sotiras@wustl.edu). }
}

\maketitle

\begin{abstract}
  Multiple Instance Learning (MIL) has been widely used in weakly supervised whole slide image (WSI) classification. Typical MIL methods include a feature embedding part, which embeds the instances into features via a pre-trained feature extractor, and an MIL aggregator that combines instance embeddings into predictions. Most efforts have typically focused on improving these parts. This involves refining the feature embeddings through self-supervised pre-training as well as modeling the correlations between instances separately. 
  In this paper, we proposed a sparsely coding MIL (SC-MIL) method that addresses those two aspects at the same time by leveraging sparse dictionary learning. The sparse dictionary learning captures the similarities of instances by expressing them as sparse linear combinations of atoms in an over-complete dictionary. In addition, imposing sparsity improves instance feature embeddings by suppressing irrelevant instances while retaining the most relevant ones. To make the conventional sparse coding algorithm compatible with deep learning, we unrolled it into a sparsely coded module leveraging deep unrolling. The proposed SC module can be incorporated into any existing MIL framework in a plug-and-play manner with an acceptable computational cost. The experimental results on multiple datasets demonstrated that the proposed SC module could substantially boost the performance of state-of-the-art MIL methods. 
  

\end{abstract}


\section{Introduction}
\label{sec:introduction}
\IEEEPARstart{T}{he} gigapixel resolution of digital whole slide images (WSIs) enables viewing and analyzing the entire tissue sample in a single image. However,  the size and complexity of the images pose significant challenges for pathologists~\cite{he2012histology}. As a consequence, there is increasing demand for automated workflows to assist in WSI diagnosis.
This has propelled the adoption and development of deep learning-based methods for WSI classification~\cite{zhou2020comprehensive,kather2019deep,coudray2018classification,ilse2018attention,li2021dual,shao2021transmil,zhang2022dtfd}. 
However, applying traditional deep learning methods to WSI classification is challenging due to the gigapixel resolution of WSIs and the absence of pixel-level annotations~\cite{litjens2016deep}. Weakly-supervised multiple instance learning (MIL)~\cite{ilse2018attention,li2021dual,shao2021transmil,zhang2022dtfd} has been proposed to address the aforementioned challenges by only leveraging image-level annotations.

In the application of MIL for WSI classification, each WSI is treated as a bag consisting of non-overlapping patches cropped from the WSI slide, with each patch serving as an unlabeled instance. The bag is labeled as positive if at least one of the instances exhibits disease; otherwise, it is labeled as negative. In the context of WSIs, MIL is commonly implemented using a two-stage approach. First, the cropped patches are converted into feature embeddings through a fixed feature extractor. A fixed extractor is preferred over a learned one due to the prohibitively expensive computation for back-propagating with thousands of instances in a bag. Second, an MIL aggregator is applied to combine the local instance feature embeddings to make bag predictions. Such a two-stage learning scheme is potentially suboptimal because the noisy feature embeddings and imbalanced instances (i.e., the positive instances make up only a small portion of all patches in a positive bag) may trap the MIL aggregator into learning an erroneous mapping between embeddings and labels. Besides, the weak supervisory signal hinders the MIL aggregator from capturing correlations between instances~\cite{shao2021transmil,xiang2023exploring,li2021dual}. 

Previous attempts at MIL tackled these two challenges separately. The first class of methods focused on refining the extracted feature embeddings by leveraging self-supervised pretraining~\cite{li2021dual, lu2019semi, xiang2023exploring, li2023task}. However, these methods require large data and an additional computationally expensive training stage. 
The second class of methods focused on improving the MIL aggregator, so that it can better capture cross-instance correlations as well as global representations of positive/negative instances~\cite{lu2021data, zhao2020predicting,li2021dual,zhang2022dtfd,xiang2023exploring}. Despite their seemingly distinct approaches, we argue that these two classes of methods are strongly interrelated. This is because better instance feature embeddings that are robust and capable of modeling the invariance of the same type of biological tissues would also simplify the task of the MIL aggregator. 


In this paper, we propose to bridge the gap between
refining the feature embeddings and enhancing the MIL aggregator through a simple but effective sparse dictionary learning (SDL). For this purpose, the feature embeddings of instances are expressed as a linear combination of atoms in an overcomplete dictionary. Accordingly, positive/negative instances from the same tissue type should be represented by similar combinations of atoms, which naturally capture the global representations of instances. The over-complete dictionary offers flexibility to model the variability among instances with the same tissue type. In addition, the inherent sparsity of SDL leads to compact and robust representations that better characterize instances, enhancing the initial noisy feature embeddings.

\subsection{Related Work}
\subsubsection{MIL in WSI classification}
MIL methods can be broadly divided into two major categories: instance-level MIL and bag-level MIL. Typically, the instance-level methods~\cite{campanella2019clinical, feng2017deep, hou2016patch, lerousseau2020weakly, xu2019camel}  start with training a network to predict instance-level labels that are assigned by propagating the bag-level label to each of its instances. Afterward, they aggregate the predicted instance-level labels to obtain the corresponding bag-level label. However, due to the fact that only a small fraction of positive instances in a bag are associated with a disease in WSIs,
the negative instances in a positive bag are often mislabeled. Despite numerous attempts to purify the instance-level labels, empirical studies have consistently shown that instance-level methods exhibit inferior performance compared to their bag-level counterparts~\cite{shao2021transmil, wang2018revisiting}. 

Bag-level MIL methods~\cite{ilse2018attention, li2021dual, shao2021transmil, zhang2022dtfd, wang2018revisiting, lu2021data, sharma2021cluster, zhu2017deep, lu2019semi, xiang2023exploring, li2023task, zhao2020predicting} follow a two-stage learning process: they first embed the instances into feature representation using a pre-trained feature extractor and then perform MIL aggregation to generate bag-level predictions. Previous explorations in bag-level MIL primarily focused on two directions. The first direction is to enhance the MIL aggregator. Following this direction of work, the attention-based MIL~\cite{ilse2018attention, lu2021data} converted the traditional non-parametric poolings, e.g., max/mean-pooling~\cite{wang2018revisiting}, into trainable ones through an attention mechanism. However, initial attempts treated each instance independently without considering their similarities. Follow-up works addressed this limitation by leveraging graph convolutional networks~\cite{zhao2020predicting}, non-local attention~\cite{li2021dual}, transformer~\cite{shao2021transmil}, and knowledge distillation~\cite{zhang2022dtfd}.
The second direction is to improve the feature embedding by leveraging self-supervised pre-training~\cite{li2021dual, lu2019semi, xiang2023exploring, li2023task}. However, these methods require a large amount of data for task-specific training and are computationally expensive. 

We approached the problem in a novel way that introduces sparse coding into MIL. 
Although our work shares some limited similarity with iterative low-rank attention (ILRA-MIL)~\cite{xiang2023exploring} in leveraging low-rank properties of instances, it is fundamentally different in several key aspects: \textbf{(i)} SDL is more flexible and adaptable than low-rank projection, as the learned over-complete dictionary can represent more complex, diverse, and irregular patterns of instances than a low-rank matrix. \textbf{(ii)} The sparsity in SDL leads to more compact and robust representations than dense representations provided by a low-rank projection. \textbf{(iii)} ILRA still treats feature enhancements and the MIL aggregator as two separate components: a low-rank guided self-supervised pre-training mechanism and a low-rank guided attention mechanism. In contrast, the proposed method offers a unified module for enhancing feature embeddings and MIL aggregation. \textbf{(iv)} ILRA is tailored to the transformer-based MIL aggregator. Whereas, the proposed method can be easily plugged into existing MIL frameworks without changing the network architectures of their respective aggregators.


\subsubsection{Sparse Dictionary Learning and Algorithmic Unrolling} 
Sparse dictionary learning is widely used in the realm of machine learning and signal processing, with applications in image restoration~\cite{elad2006image, gu2015convolutional, xiao2023sc,wang2015deep}, image classification~\cite{li2022revisiting}, and compressed sensing~\cite{zhang2018ista}. Its objective is to obtain a compact and robust representation of data through a sparse linear combination of atoms in a dictionary that can effectively characterize the input signal~\cite{elad2006image, yang2010image}. This process is formulated as an optimization problem solved by iterative algorithms, such as K-SVD~\cite{aharon2006k}, iterative shrinkage-thresholding (ISTA)~\cite{daubechies2004iterative}, and fast ISTA (FISTA)~\cite{beck2009fast}. However, these iterative algorithms are not directly compatible with deep neural networks through a task-specific end-to-end optimization~\cite{bahrampour2015multimodal,mairal2011task}. Additionally, the convergence of the iterative algorithms is highly sensitive to the choice of hyperparameters, e.g., stepsize and the strength of sparsity regularization. The algorithmic unrolling~\cite{gregor2010learning,scetbon2021deep, xiao2023sc} addressed these problems by reformulating sparse coding as layers in network architectures that can be directly optimized for certain tasks through backpropagation. 

In this work, we consider the unrolled version of a learnable ISTA for sparse coding, which is related to the one used in~\cite{scetbon2021deep, xiao2023sc}, but differs from them in two main aspects: \textbf{(i)} The works in~\cite{scetbon2021deep, xiao2023sc} were designed for traditional image reconstruction tasks. Consequently, they performed matrix multiplication using the learned dictionary and sparse coefficients to reconstruct the original images. In contrast, the proposed method directly uses the learned sparse coefficients to represent the instances. The dictionary is maintained to capture global representations of instances. This is because the sparse coefficients capture the most relevant information for instances while suppressing irrelevant background information. \textbf{(ii)} Unlike~\cite{scetbon2021deep, xiao2023sc} where only one sparsity regularization strength is learned for each image, Our method learns separate sparsity strength for each instance within a bag. This is because each instance may contribute differently to representing a bag (see \Cref{sec:method:c} for details). 

\subsection{Contribution}

The main contribution of this paper is the introduction of sparse coding (SC) to multiple instance learning to refine both feature embeddings and model instance correlations as well as variability. 
However, the conventional algorithms for SDL are not directly compatible with deep neural networks to learn task-specific sparse coding and require extensive hyperparameter tuning. 
We designed an unrolled SC module for sparse dictionary learning that can be optimized by training the MIL task in an end-to-end manner. The proposed SC module is orthogonal to existing MIL frameworks and can be easily integrated into them in a plug-and-play fashion with acceptable additional computational cost. The experimental results on multiple datasets and various tasks demonstrated the effectiveness of the proposed method in boosting the performance of recent state-of-the-art MIL methods. 

\begin{figure*}[!t]
    \centering
    \includegraphics[width=\textwidth]{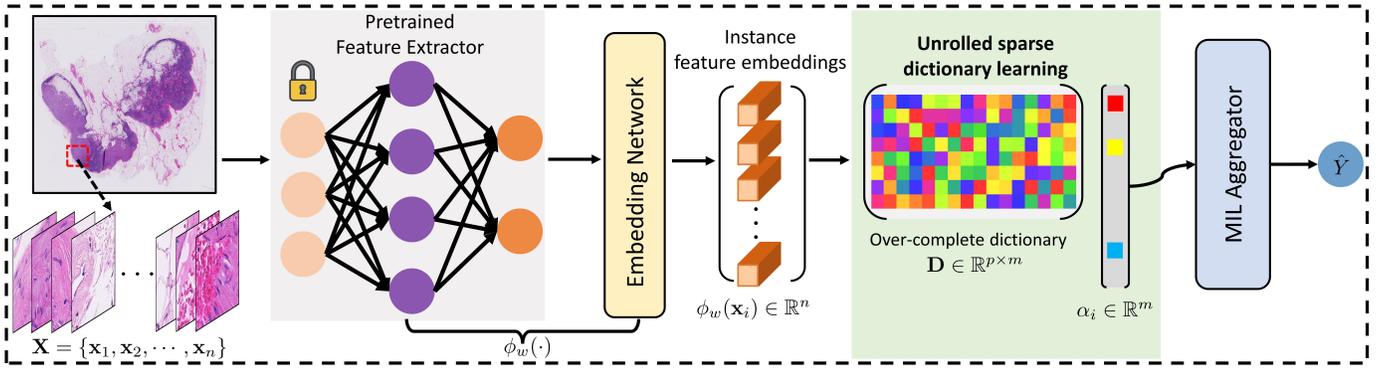}
    \caption{The workflow of the proposed SC-MIL framework in WSI classification. The sparse coding (SC) module conducts end-to-end unrolled sparse dictionary learning and can be easily integrated into any multiple instance learning (MIL) framework in a plug-and-play fashion.}
    \label{fig:mil}
\end{figure*}

\section{Method}
In this section, we first introduce the standard MIL formulation (\Cref{sec:formulation}) and then discuss the integration of sparse coding into a standard MIL framework (\Cref{sec:scmil}). The yielded sparse coding MIL framework is depicted in~\Cref{fig:mil}. Finally, we discuss how to design and learn a task-specific sparse coding for MIL in an end-to-end fashion by leveraging the algorithmic unrolling (\Cref{sec:method:c}).

\subsection{Problem Formulation}\label{sec:formulation}
Without loss of generality, let us consider the problem of bag-level binary MIL classification. Its objective is to learn a mapping from a bag of instances $\textbf{X} = \{\textbf{x}_1, \textbf{x}_2, \cdots, \textbf{x}_n \}$ to bag-level corresponding label $Y \in \{0, 1\}$. In the context of WSI classification, each bag $\textbf{X}$ denotes a WSI with $n$ tiled patches, where $n$ may vary from bag to bag. Mathematically, the bag-level binary MIL classification is defined as:
\begin{equation}\nonumber
    \begin{split}
        Y = \begin{cases}
            0, \ \text{iff} \ \sum_{i=1}^{n} y_{i} = 0 \\
            1, \ \text{otherwise},
        \end{cases}
    \end{split}
\end{equation}
where $y_{i} \in \{0, 1\}$ denotes the unknown instance-level label of the $i$-th instance. The instance-level labels are, however, unavailable in most scenarios. 

A standard deep learning MIL framework contains three main components. First, instances are embedded into feature vectors via a pretrained feature extractor network $\phi_{w}$ parameterized by $w$ (e.g., a ResNet~\cite{resnet}) and a simple trainable embedding network (e.g., a fully-connected layer). Second, the instance feature embeddings are then aggregated by an MIL aggregator $\sigma$, where $\sigma$ is a permutation-invariant function. Third, a bag-level classifier $f_{\text{cls}}$ is applied to the aggregated features by an MIL aggregator to produce the bag-level probability prediction $\hat{Y} \in [0, 1]$:
\begin{equation}\nonumber
    \hat{Y} =  f_{cls}\left( \sigma (\{ \phi_{w}(\textbf{x}_1), \phi_{w}(\textbf{x}_2), \cdots, \phi_{w}(\textbf{x}_n) \}) \right). 
\end{equation}

\subsection{Sparse Coding MIL}\label{sec:scmil}
The proposed sparsely coded MIL is constructed by plugging the proposed SC module at the very beginning of the MIL aggregator (see \Cref{fig:mil}).
Specifically, we assume that the initial instance feature embeddings $\phi_w(\textbf{x}_i)$ can be expressed as a linear combination of $s \ll m$ atoms from an over-complete dictionary $\textbf{D} \in \mathbb{R}^{p \times m}$, where $m$ and $p$ are the number of atoms and dimension of each atom in a dictionary, respectively. 
Mathematically, an instance can be expressed as $\phi_w(\textbf{x}_i)=\textbf{D}\alpha_i$, where $\alpha_i \in \mathbb{R}^{m}$ denotes the sparse coefficients for each instance embedding $\phi_w(\textbf{x}_i)$. This process is formally defined as sparse dictionary learning by optimizing the following objective function:
\begin{equation} \label{eqn:sc}
    \begin{split}
        \hat{\alpha_i} = \argmin_{\alpha_i} \frac{1}{2} || \textbf{D}  \alpha_i - \phi_w(\textbf{x}_i)  ||_2^2 + \lambda ||\alpha_i||_1, \ \lambda > 0
    \end{split}
\end{equation}
where $\lambda$ controls the strength of the sparsity regularization. The $\ell_1$ regularization of the $\alpha_i$ is an approximate relaxation of the $\ell_0$ sparsity, which results in a convex optimization. 
An effective solver of \Cref{eqn:sc} is the Iterative Soft Thresholding Algorithm (ISTA)~\cite{daubechies2004iterative}, which is given as a proximal update:
\begin{equation} \label{eqn:ista}
    \begin{split}
        \hat{\alpha_{i}}^{(t+1)} &=
        S_{\lambda} \left(\hat{\alpha_i}^{(t)} - \frac{1}{\mu} \textbf{D}^{T} (\textbf{D} \hat{\alpha_{i}}^{(t)} - \phi_w(\textbf{x}_i) ) \right) \\  & \quad \quad \quad \quad \quad \text{with} \ \hat{\alpha_i}^{(0)} = 0, 
    \end{split}
\end{equation}
where $\mu$ is the stepsize, and $t$ denotes $t$-th iteration. $S_{\lambda}(\cdot)$ is the element-wise soft-thresholding operator, serving as the proximal projection:
\begin{equation}
   [S_{\lambda}(\textbf{v})]_j = sign(v_j) \cdot \operatorname{max}\{|v_j| - \lambda, 0\},
\end{equation}
where $v_j$ is the $j$-th element of $\mathbf{v}$, and $\operatorname{max}\{\cdot \}$ is the element-wise max operator. It is worth noting that we employ learnable ISTA~\cite{gregor2010learning}, where both the dictionary $\textbf{D}$ and sparse coefficients $\alpha_i$ are learned.

\begin{figure*}[!t]
    \centering
    \includegraphics[width=0.9\textwidth]{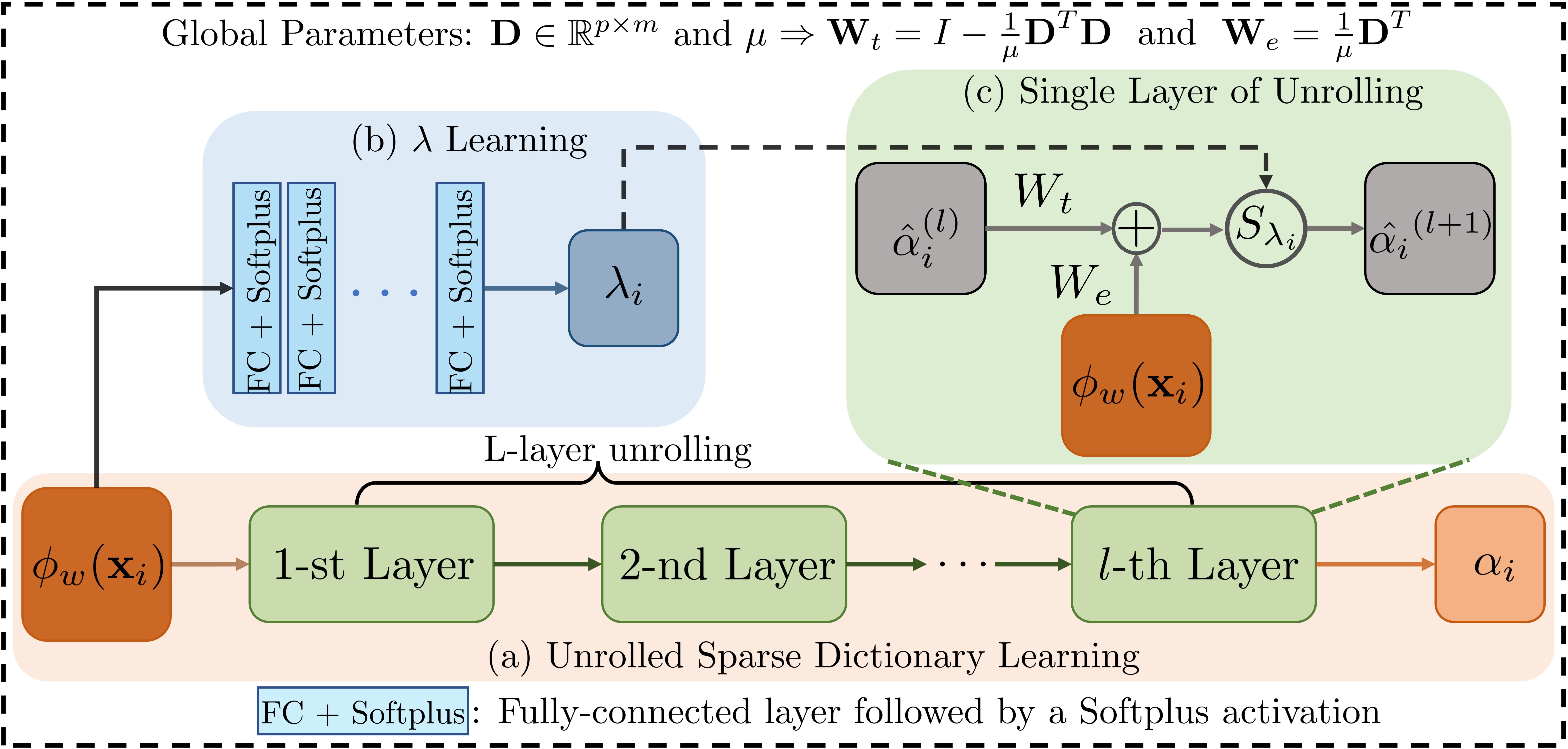}
    \caption{The proposed SC module: (a) The unrolled ISTA learning scheme of the sparse dictionary learning; (b) The $\lambda$ learning module, which is implemented as a feed-forward network; (c) A single network layer of the unrolling network for sparse dictionary learning. }
    \label{fig:sc}
\end{figure*}

\subsection{Unrolled Sparse Dictionary Learning}\label{sec:method:c}
Even though the ISTA defined in~\Cref{eqn:ista} is block-wise convex, its convergence requires a proper choice of stepsize $\mu$ and regularization strength $\lambda$. Furthermore, the dictionary should be optimized for a specific task given the input data, instead of a predefined one in many traditional SDL~\cite{mairal2011task,qayyum2015designing,bahrampour2015multimodal}. 
Leveraging the proximal operator, the ISTA-based sparse dictionary learning can be unrolled into a fully learnable scheme. Specifically, the reparameterization of \Cref{eqn:ista} yields the learnable ISTA (LISTA):
\begin{equation}
    \begin{split}
       \hat{\alpha_i}^{(t+1)} &= S_{\lambda} \left( \textbf{W}_t \hat{\alpha_i}^{(t)} + \textbf{W}_e \phi_w(\textbf{x}_i)  \right) \ \text{with} \\
       \textbf{W}_t &= I - \frac{1}{\mu} \textbf{D}^{T} \textbf{D} \ \ \text{and} \ \ \textbf{W}_e = \frac{1}{\mu} \textbf{D}^{T},
    \end{split}
    \label{eqn:LISTA}
\end{equation}
where the parameters (i.e., $\textbf{D}$, $\mu$, and $\lambda$) can be optimized in a trainable fashion. Given the dictionary $\textbf{D}$, stepsize $\mu$, and sparsity strength $\lambda$, the update rule in \Cref{eqn:LISTA} can be recast into a single network layer, as depicted in \Cref{fig:sc}(c). 
The cascaded repeat of such a single network layer $L$ times results in an L-layer unrolled network for sparse dictionary learning (see \Cref{fig:sc}(a)), while maintaining the same computational path as \Cref{eqn:ista}. This is also equivalent to performing $L$ iterations of LISTA update outlined in \Cref{eqn:LISTA}.  Accordingly, we unrolled the ISTA-based sparse dictionary learning into a single module called the SC module.
We would like to point out that the number of unrolled layers $L$ is a hyperparameter that can be tuned to balance the trade-off between model complexity and performance. 

We then discuss how to learn the three key components (i.e., dictionary $\textbf{D}$, sparsity strength $\lambda$, and stepsize $\mu$) in the proposed unrolled SDL.
\subsubsection{Learning the over-complete dictionary}
The over-complete dictionary $\textbf{D}$ in the sparse dictionary learning can be used to model the similarity and variability among instances, which is a key requirement for solving the MIL problem. To achieve a globally invariant representation for instances belonging to the same tissue type, we set the dictionary $\textbf{D}$ as a global parameter that is optimized across all bags/WSIs. As each operation (i.e., matrix multiplication, summation, soft-thresholding) in a single unrolling layer is differentiable, the optimization of $\textbf{D}$ can be achieved by backpropagation via training the binary MIL classification task in an end-to-end fashion. To speed up its convergence, the dictionary is initialized with an over-complete discrete cosine transform matrix~\cite{qayyum2015designing}.  

\subsubsection{Learning the optimal $\lambda$}
The strength of the sparsity regularization $\lambda$ is an important parameter to select in the standard ISTA update outlined in \Cref{eqn:ista}. The value of $\lambda$ determines a trade-off between the sparsity and expressiveness of the sparse dictionary learning and therefore requires careful tuning. However, within the context of MIL, the optimal choice of $\lambda$ can vary from bag to bag, making it challenging to tune manually. Inspired by~\cite{scetbon2021deep}, we formulated the estimation of the optimal $\lambda_i$ for each instance as a regression task. Specifically, $\lambda_i$ was parameterized as a simple feed-forward network (FFN) $f_{\theta}(\phi_w(\textbf{x}_i))$ (see~\Cref{fig:sc}(b)), where $\theta$ denotes the parameters of the FFN. In this work, the FFN consisted of three fully-connected layers, each followed by a Softplus activation~\cite{glorot2011deep}. 

We would like to point out two main differences in the design of the proposed SC compared to~\cite{scetbon2021deep,xiao2023sc}: \textbf{(i)} Unlike in~\cite{scetbon2021deep} and \cite{xiao2023sc}, where only one $\lambda$ was learned for each image, we learned $n$ $\lambda$s, one for each instance within a single WSI image. This distinction arises from the assumption in ~\cite{scetbon2021deep} and \cite{xiao2023sc} that patches within each image should contribute equally to image reconstruction tasks.  In contrast, our approach acknowledges that instances contribute differently to the MIL classification task. \textbf{(ii)} We used a Softplus activation function, instead of the rectified linear unit (ReLU) activation used in~\cite{scetbon2021deep,xiao2023sc} to learn $\lambda$. As described in~\Cref{eqn:sc}, $\lambda$ is constrained to be a positive value. However, we observed that the ReLU activation caused instability in SDL optimization in our initial experiments. This is because of the discontinuity of the gradient of ReLU. Accordingly, we used Softplus activation, a smooth approximation of ReLU, to alleviate this problem by slightly relaxing the constraint $\lambda > 0$. 

\subsubsection{Learning the optimal stepsize $\mu$}
The choice of stepsize $\mu$ is another key factor affecting the convergence of the ISTA. One effective choice that is commonly used is to set $\mu$ as the square of the spectral norm of the dictionary~\cite{scetbon2021deep}. Nonetheless, the optimal stepsize is prone to vary across different datasets and tasks. To determine the optimal stepsize $\mu$ for a given dataset, we made it learnable by setting it as a global parameter initialized with the square of the spectral norm of the dictionary.

\section{Experiments and Results}
We conducted several experiments on multiple datasets, including five classical MIL benchmarks, the CAMELYON16 dataset~\cite{bejnordi2017diagnostic}, and the Cancer Genome Atlas non-small cell lung cancer (TCGA-NSCLC) dataset to validate the effectiveness of the proposed method.

\begin{table*}[!t]
\caption{Performance on CAMELYON16 and TCGA-NSCLC datasets using features extracted by \textbf{ResNet-18}, \textbf{Swin-ViT}, and \textbf{CTransPath}. The mean ($\pm$ standard deviation) of classification accuracy (\%) and AUC (\%) were reported. The +SC denotes incorporating the proposed SC module into the corresponding MIL methods. $\Delta$ denotes the performance difference, with \textcolor{blue!55}{blue} indicating gain and \textcolor{Gray!}{gray} indicating loss. Integrating the SC module led to improved performance across multiple methods and datasets for WSI classification. ($^*: p < 0.05$, with Wilcoxon signed-rank test to the corresponding method without SC.)}
\centering
\resizebox{0.8\textwidth}{!}{
\begin {tabular}{p{0.9cm}p{2.8cm}p{1.8cm}cccc}
\toprule %
\multicolumn{3}{c}{\backslashbox[50mm]{\textbf{Method}}{\textbf{Performance}}} & \multicolumn{2}{c}{CAMELYON16} & \multicolumn{2}{c}{TCGA-NSCLC} \\
 \cmidrule(r){4-5} \cmidrule(r){6-7}
&  & & Accuracy & AUC & Accuracy & AUC \\
\toprule %
\multirow{3}{*}[-1.2cm]{\rotatebox[origin=c]{90}{\makecell{\textbf{ResNet-18} \\ ImageNet \ Pretrained}}} & ABMIL-Gated & \makecell{\\+SC \\ $\Delta$} & \makecell{$83.41{\pm1.2}$ \\ $87.75{\pm1.4}$ \\  \color{blue!55}+ 4.34} & \makecell{$85.26{\pm0.7}$ \\ $90.36{\pm1.2}$ \\ \color{blue!55}+5.10} & \makecell{$85.32{\pm1.8}$ \\ $87.72{\pm1.7}$ \\ \color{blue!55}+2.40} & \makecell{$91.25{\pm1.1}$ \\ $93.46{\pm1.2}$ \\ \color{blue!55}+2.21}\\ 
\cmidrule (l ){2 -7}
& DSMIL & \makecell{\\+SC \\ $\Delta$} & \makecell{$84.03{\pm 2.2}$ \\ $88.06{\pm 1.9}$  \\ \color{blue!55}+4.03} & \makecell{$87.52{\pm 1.3}$  \\ $92.48{\pm 1.1}$  \\ \color{blue!55}+4.96} & \makecell{$84.07{\pm 1.7}$  \\ $86.76{\pm 2.3}$  \\ \color{blue!55}+2.69} & \makecell{$92.07{\pm 2.4}$  \\ $93.44{\pm 1.2}$  \\ \color{blue!55}+1.37}\\ \cmidrule (l ){2 -7}
& TransMIL & \makecell{\\+SC \\ $\Delta$} &  \makecell{$86.67{\pm 2.0}$ \\ $88.22{\pm 1.1}$ \\ \color{blue!55}+1.55}  &  \makecell{$90.64{\pm 2.0}$ \\ $91.38{\pm 0.6}$ \\ \color{blue!55}+0.74}  &  \makecell{$87.81{\pm 1.2}$ \\ $89.35{\pm 1.0}$ \\ \color{blue!55}+1.54} &  \makecell{$94.53{\pm 0.9}$ \\ $94.86{\pm 0.9}$ \\ \color{blue!55}+0.33}\\ \cmidrule (l ){2 -7}
& DTFD-MIL (MaxMinS) & \makecell{\\+SC \\ $\Delta$} & \makecell{$86.05{\pm 1.8}$ \\ $87.44{\pm 2.0}$ \\ \color{blue!55}+1.39} & \makecell{$90.97{\pm 0.4}$ \\ $91.14{\pm 1.3}$ \\ \color{blue!55}+0.17} & \makecell{$87.53{\pm 1.8}$ \\ $89.44{\pm 1.6}$ \\ \color{blue!55}+1.91} & \makecell{$93.37{\pm 1.1}$ \\ $94.50{\pm 0.7}$ \\ \color{blue!55}+1.13}\\ \cmidrule (l ){2 -7}
& DTFD-MIL (AFS) & \makecell{\\+SC \\ $\Delta$} & \makecell{$87.75{\pm 1.3}$ \\ $87.91{\pm 0.6}$ \\ \color{blue!55}+0.16} & \makecell{$89.16{\pm 0.4}$ \\ $91.35{\pm 0.8}$ \\ \color{blue!55}+2.19} & \makecell{$88.49{\pm 1.1}$ \\ $89.92{\pm 0.8}$ \\ \color{blue!55}+1.43} & \makecell{$94.31{\pm 1.3}$ \\ $94.84{\pm 1.3}$ \\ \color{blue!55}+0.53}\\
\midrule
\multirow{3}{*}[-1.2cm]{\rotatebox[origin=c]{90}{\makecell{\textbf{Swin-ViT} \\ ImageNet \ Pretrained}}} & ABMIL-Gated & \makecell{\\+SC \\ $\Delta$} & \makecell{$83.41{\pm 1.3}$ \\ $86.98{\pm 1.6}$ \\ \color{blue!55}+3.57} & \makecell{$85.44{\pm 0.8}$ \\ $90.99{\pm 0.7}$ \\ \color{blue!55}+5.55} & \makecell{$88.20{\pm 0.5}$ \\ $90.12{\pm 0.5}$ \\ \color{blue!55}+1.92} & \makecell{$94.24{\pm 0.7}$ \\ $96.10{\pm 0.6}$ \\ \color{blue!55}+1.86}\\ 
\cmidrule (l ){2 -7}
& DSMIL & \makecell{\\ +SC \\ $\Delta$} & \makecell{$85.74{\pm 0.6}$ \\ $88.22{\pm 1.0}$ \\ \color{blue!55}+2.48} & \makecell{$88.20{\pm 0.7}$ \\ $91.04{\pm 0.7}$ \\ \color{blue!55}+2.84} & \makecell{$84.55{\pm 1.4}$ \\ $87.91{\pm 2.3}$ \\ \color{blue!55}+3.36} & \makecell{$93.43{\pm 1.3}$ \\ $94.91{\pm 1.1}$ \\ \color{blue!55}+1.48}\\ \cmidrule (l){2 -7}
& TransMIL & \makecell{\\+SC \\ $\Delta$} & \makecell{$87.13{\pm 1.5}$ \\ $87.91{\pm 1.2}$ \\ \color{blue!55}+0.78} & \makecell{$90.99{\pm 0.6}$ \\ $92.05{\pm 1.0}$ \\ \color{blue!55}+1.06} & \makecell{$90.79{\pm 0.6}$ \\ $92.03{\pm 1.6}$ \\ \color{blue!55}+1.24} & \makecell{$96.05{\pm 0.6}$ \\ $96.57{\pm 1.0}$ \\ \color{blue!55}+0.52}\\ 
\cmidrule (l ){2 -7}
& DTFD-MIL (MaxMinS) & \makecell{\\+SC \\ $\Delta$} & \makecell{$87.44{\pm 0.8}$ \\ $90.39{\pm 1.4}$ \\ \color{blue!55}+2.94} & \makecell{$90.51{\pm 0.6}$ \\ $93.10{\pm 1.5}$ \\ \color{blue!55}+2.59} & \makecell{$90.02{\pm 1.6}$ \\ $92.13{\pm 1.2}$ \\ \color{blue!55}+2.11} & \makecell{$94.48{\pm 0.4}$ \\ $95.95{\pm 0.4}$ \\ \color{blue!55}+1.47}\\ \cmidrule (l ){2 -7}
& DTFD-MIL (AFS) & \makecell{\\+SC \\ $\Delta$} & \makecell{$85.89{\pm 0.9}$ \\ $87.09{\pm 3.0}$ \\ \color{blue!55}+1.20} & \makecell{$87.10{\pm 0.5}$ \\ $90.49{\pm 0.8}$ \\ \color{blue!55}+3.39} & \makecell{$91.45{\pm 0.9}$ \\ $92.03{\pm 1.3}$ \\ \color{blue!55}+0.58} & \makecell{$95.72{\pm 0.6}$ \\ $96.44{\pm 0.5}$ \\ \color{blue!55}+0.72}\\ 
\midrule
\multirow{3}{*}[-1.2cm]{\rotatebox[origin=c]{90}{\makecell{\textbf{CTransPath} \\ Self-supervised \ Pretrained}}} & ABMIL-Gated & \makecell{\\+SC \\ $\Delta$} & \makecell{$95.50{\pm 0.3}$ \\ $96.59{\pm 0.8}$ \\ \color{blue!55}+1.09} & \makecell{$95.62{\pm 0.1}$ \\ $98.62{\pm 0.5}$ \\ \color{blue!55}+3.00} &
\makecell{$90.76{\pm 1.3}$ \\ $93.38{\pm 1.0}$ \\ \color{blue!55}+2.62} & \makecell{$95.93{\pm 0.8}$ \\ $97.34{\pm 0.4}$ \\ \color{blue!55}+1.41}\\ 
\cmidrule (l ){2 -7}
& DSMIL & \makecell{\\ +SC \\ $\Delta$} & \makecell{$94.73{\pm 1.7}$ \\ $95.97{\pm 0.9}$ \\ \color{blue!55}+1.24} & \makecell{$95.01{\pm 0.6}$ \\ $98.35{\pm 1.1}$ \\ \color{blue!55}+3.34} & \makecell{$89.35{\pm 0.7}$ \\ $92.13{\pm 0.8}$ \\ \color{blue!55}+2.78} & \makecell{$96.43{\pm 0.5}$ \\ $97.72{\pm 0.5}$ \\ \color{blue!55}+1.29}\\ \cmidrule (l ){2 -7}
& TransMIL & \makecell{\\+SC \\ $\Delta$} & \makecell{$96.43{\pm 1.1}$ \\ $96.90{\pm 0.7}$ \\ \color{blue!55}+0.47} & \makecell{$99.12{\pm 0.2}$ \\ $98.72{\pm 0.2}$ \\ \color{Gray!}-0.40} & \makecell{$93.47{\pm 0.5}$ \\ $94.63{\pm 1.6}$ \\ \color{blue!55}+1.16} & \makecell{$97.80{\pm 0.5}$ \\ $98.28{\pm 0.7}$ \\ \color{blue!55}+0.48}\\ 
\cmidrule (l ){2 -7}
& DTFD-MIL (MaxMinS) & \makecell{\\+SC \\ $\Delta$} & \makecell{$96.59{\pm 0.8}$ \\ $96.90{\pm 1.0}$ \\ \color{blue!55}+0.31} & \makecell{$98.81{\pm 0.1}$ \\ $98.88{\pm 0.2}$ \\ \color{blue!55}+0.07} & \makecell{$93.67{\pm 0.7}$ \\ $95.01{\pm 0.6}$ \\ \color{blue!55}+1.34} & \makecell{$97.46{\pm 0.4}$ \\ $97.88{\pm 0.4}$ \\ \color{blue!55}+0.42}\\ \cmidrule (l ){2 -7} 
& DTFD-MIL (AFS) & \makecell{\\+SC \\ $\Delta$} & \makecell{$96.59{\pm 1.3}$ \\ $97.36{\pm 0.8}$ \\ \color{blue!55}+0.77} & \makecell{$98.61{\pm 0.0}$ \\ $98.96{\pm 0.2}$ \\ \color{blue!55}+0.35} & \makecell{$93.28{\pm 0.8}$ \\ $94.53{\pm 1.4}$ \\ \color{blue!55}+1.25} & \makecell{$97.52{\pm 0.4}$ \\ $98.04{\pm 0.7}$ \\ \color{blue!55}+0.52}\\
\midrule
& & \makecell{Average $\Delta^*$} & \color{blue!55}+1.75$^*$ & \color{blue!55}+2.33$^*$ & \color{blue!55}+1.89$^*$ & \color{blue!55}+1.05$^*$ \\
\bottomrule
\end{tabular}}
\label{tab:wsi}
\end{table*}
\begin{table*}[!t]
  \caption{Performance on five classical MIL benchmark datasets. Each experiment was performed five times with 10-fold cross-validation. We reported the mean of the classification accuracy ($\pm$ the standard deviation of the mean). Previous benchmark results were obtained from~\cite{ilse2018attention, li2021dual} under the same experimental settings. The best performance is marked in \textbf{bold}, while the second best performance is highlighted with an \underline{underline}. Integrating the proposed SC module to ABMIL led to a significant performance gain and outperformed all baseline methods. ($^*: p < 0.05$, with Wilcoxon signed-rank test to all baseline methods.) }
  \centering
  \resizebox{0.85\textwidth}{!}{
  \begin{tabular}{l|cccccc}
    \toprule
    \backslashbox[30mm]{\textbf{Method}}{\textbf{Performance}}     & MUSK1  & MUSK2 & FOX & TIGER & ELEPHANT & Average \\
    \hline
    mi-Net~\cite{wang2018revisiting} & 0.889 $\pm$ 0.039  & 0.858 $\pm$ 0.049 & 0.613 $\pm$ 0.035 &  0.824 $\pm$ 0.034 &  0.858 $\pm$ 0.037 & 0.808   \\
    MI-Net~\cite{wang2018revisiting} &  0.887 $\pm$ 0.041 & 0.859 $\pm$ 0.046 & 0.622 $\pm$ 0.038 & 0.830 $\pm$ 0.032 & 0.862 $\pm$ 0.034 & 0.812     \\
    MI-Net with DS~\cite{wang2018revisiting} & 0.894 $\pm$ 0.042 & 0.874 $\pm$ 0.043 & 0.630 $\pm$ 0.037 & 0.845 $\pm$ 0.039 & 0.872 $\pm$ 0.032 & 0.823   \\
    MI-Net with RC~\cite{wang2018revisiting} & 0.898 $\pm$ 0.043 & 0.873 $\pm$ 0.044 & 0.619 $\pm$ 0.047 & 0.836 $\pm$ 0.037 & 0.857 $\pm$ 0.040  & 0.817 \\
    ABMIL~\cite{ilse2018attention} &  0.892 $\pm$ 0.040 & 0.858 $\pm$ 0.048 & 0.615 $\pm$ 0.043 & 0.839 $\pm$ 0.022 & 0.868 $\pm$ 0.022 & 0.814  \\
    ABMIL-Gated~\cite{ilse2018attention} & 0.900 $\pm$ 0.050 & 0.863 $\pm$ 0.042 & 0.603 $\pm$ 0.029 & 0.845 $\pm$ 0.018 & 0.857 $\pm$ 0.027 & 0.814  \\
    GNN-MIL~\cite{tu2019multiple} & 0.917 $\pm$ 0.048 & 0.892 $\pm$ 0.011 & 0.679 $\pm$ 0.007 & 0.876 $\pm$ 0.015 & 0.903 $\pm$ 0.010  & 0.853 \\
    DP-MINN~\cite{yan2018deep}  & 0.907 $\pm$ 0.036 & 0.926 $\pm$ 0.043 & 0.655 $\pm$ 0.052 & 0.897 $\pm$ 0.028 & 0.894 $\pm$ 0.030 & 0.856 \\
    \hline
    NLMIL~\cite{wang2018non} & 0.921 $\pm$ 0.017 & 0.910 $\pm$ 0.009 & 0.703 $\pm$ 0.035 & 0.857 $\pm$ 0.013 & 0.876 $\pm$ 0.011 & 0.853 \\
    ANLMIL~\cite{zhu2019asymmetric} & 0.912 $\pm$ 0.009 & 0.822 $\pm$ 0.084 & 0.643 $\pm$ 0.012 & 0.733 $\pm$ 0.068 & 0.883 $\pm$ 0.014 & 0.799 \\
    DSMIL~\cite{li2021dual} & 0.932 $\pm$ 0.023  & 0.930 $\pm$ 0.020 & 0.729 $\pm$ 0.018 & 0.869 $\pm$ 0.008 & 0.925 $\pm$ 0.007 & 0.877 \\
    \hline
    ABMIL \textbf{w/} SC & \underline{0.958 $\pm$ 0.015} & \underline{0.958 $\pm$ 0.008} & \underline{0.789 $\pm$ 0.015} & \underline{0.933 $\pm$ 0.007}  & \underline{0.949 $\pm$ 0.004} & \underline{0.917}  \\
    ABMIL-Gated \textbf{w/} SC & \textbf{{0.969 $\pm$ 0.004}} & \textbf{{0.960 $\pm$ 0.008}} & \textbf{{0.791 $\pm$ 0.007}} & \textbf{{0.948 $\pm$ 0.004}}  & \textbf{{0.956 $\pm$ 0.004}} & \textbf{0.925}$^*$   \\
    \bottomrule
  \end{tabular}
  }
  \label{tab:classic}
\end{table*}

\subsection{Dataset}
\subsubsection{Classical MIL benchmarks}
 The five classical MIL benchmark datasets consist of MUSK1, MUSK2, FOX, TIGER, and ELEPHANT datasets. The MUSK1 and MUSK2 datasets are used to predict the impact of drugs given the molecule conformations. Each bag consists of several conformations of the same molecule. 
The label for the bag is positive if at least one of its conformations has the desired drug effect, and negative, if none is effective~\cite{dietterich1997solving}. The FOX, TIGER, and ELEPHANT datasets identify if the target animal is presented in a bag. Each bag consists of a set of features extracted from segments of an image. Positive bags refer to images that contain the animals of interest, whereas negative bags are images where no such animal is present~\cite{andrews2002support}. 

\subsubsection{CAMELYON16 dataset} 
The CAMELYON16~\cite{bejnordi2017diagnostic} is a publicly available WSI dataset for detecting metastatic breast cancer in lymph node tissue. The dataset consists of 399 WSIs (one corrupted sample was discarded) of lymph node tissue, officially split into a training set of 270 samples and a testing set of 129 samples. Each WSI is accompanied by a binary label, annotated by pathologists, indicating the presence or absence of metastatic cancer in the lymph node tissue. The dataset also includes detailed annotations of the regions of interest within each WSI that contains cancerous tissue. By following the preprocessing procedures outlined in~\cite{lu2021data}, we cropped the WSIs into non-overlapping patches of size $256 \times 256$. This resulted in
a total of around 4.61 million patches at $\times20$ magnification, with an average of $11,555$ patches per bag.

\subsubsection{TCGA-NSCLC dataset} 
The TCAGA-NSCLC is a different WSI dataset that is used for identifying two sub-types of lung cancer: lung squamous cell carcinoma and lung adenocarcinoma. Following~\cite{li2021dual}, we used a total of $1,037$ diagnostic WSIs in our experiments. After performing the preprocessing outlined in~\cite{lu2021data}, roughly 13.83 million patches were extracted at a $\times20$ magnification level. On average, each bag consisted of $13,335$ patches.

\subsection{Feature Extraction}
The five classical MIL benchmarks comprise pre-extracted feature vectors of instances. A simple feed-forward network with the same architecture as in~\cite{wang2018revisiting, ilse2018attention} was deployed for further feature embedding.   
In the case of the WSI datasets, following~\cite{lin2023interventional}, we adopted three different feature extractors with different training paradigms to thoroughly evaluate the proposed method. Specifically, we chose \textbf{(i)} a ResNet-18~\cite{resnet} feature extractor pretrained using natural images (i.e., ImageNet); \textbf{(ii)} a Swin vision transformer (Swin-ViT)~\cite{liu2021swin} pretrained on ImageNet; \textbf{(iii)} a Swin-ViT pretrained on large-scale histopathological datasets using self-supervised learning (CTransPath~\cite{wang2022transformer}). For the first two feature extractors, we adopted the pretrained weights provided by \texttt{PyTorch}. For the CTransPath, we adopted the pretrained weights provided by the authors of~\cite{wang2022transformer}.


\begin{table*}[!t]
\begin{minipage}{0.625\textwidth}
\captionof{table}{Ablation studies on two key hyper-parameters (i.e., number of unrolled layer $L$ and number of atoms $m$) in the proposed SC module using ABMIL-Gated aggregator and features extracted by a ResNet-18 on the CAMELYON16 dataset. FLOPs were measured based on a bag containing 120 instances.}
   \begin{subtable}[t]{0.58\textwidth}
   \centering
   \resizebox{1.0\textwidth}{!}{
     \begin{tabular}[t]{ccc}
        \toprule
        \# Atoms ($m$) & \# Params / FLOPs & AUC \\
        \hline
        $m=64$ & 73.81K / 10.02K & 88.06 \\
        $m=128$ & 94.04K / 18.59K & 89.22 \\
        $m=256$ & 189.97K / 86.13K & 90.36 \\
        $m=512$ & 561.68K / 888.60K & 90.45 \\
        \bottomrule
       \end{tabular}}
       \caption{The number of atoms when $L=5$}
       \label{tab:atoms}
   \end{subtable}
   \hspace{\fill} \hspace{\fill}
    \begin{subtable}[t]{0.45\textwidth}
   \centering
   \resizebox{1.0\textwidth}{!}{
    \begin{tabular}[t]{ccc}
        \toprule
        \# Layers ($L$) &  FLOPs & AUC \\
        \hline
        $L=1$ & 53.93K & 88.98 \\
        $L=3$ & 70.03K & 89.38 \\
        $L=5$ & 86.13K & 90.36 \\
        $L=7$ & 102.22K & 90.24 \\
        $L=9$ & 118.31K & 90.26 \\
        \bottomrule
        \end{tabular}}
        \caption{The number of layers when $m=256$}
        \label{tab:layers}
    \end{subtable}
    \end{minipage}
    \hfill
    \begin{minipage}{0.31\textwidth}
   \centering
    \includegraphics[width=0.98\textwidth]{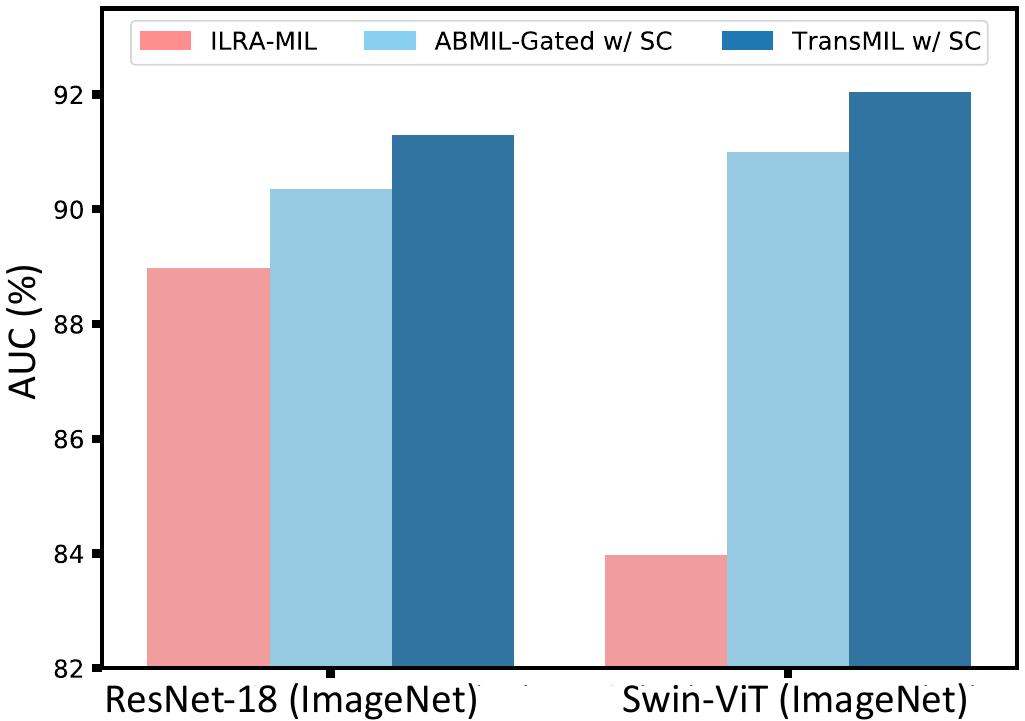}
    \captionof{figure}{Comparison between sparse coding and low-rank projection (ILRA).}
    \label{fig:sc_ilra}
    \end{minipage}
\end{table*}

\subsection{Experimental Designs}
\subsubsection{Baseline}
For the classical MIL benchmark datasets, we compared the proposed method to a series of deep learning-based MIL methods, including mi-Net and MI-Net~\cite{wang2018revisiting}, ABMIL and ABMIL-Gated~\cite{ilse2018attention}, GNN-MIL~\cite{tu2019multiple}, DP-MINN~\cite{yan2018deep}, and three non-local MIL pooling methods (i.e., NLMIL~\cite{wang2018non}, ANLMIL~\cite{zhu2019asymmetric}, and  DSMIL~\cite{li2021dual}). In the case of WSI classification, we considered plugging the proposed SC module into four recent state-of-the-art MIL aggregators, i.e., ABMIL with gated attention~\cite{ilse2018attention}, DSMIL~\cite{li2021dual}, TransMIL~\cite{shao2021transmil}, and DTFD-MIL~\cite{zhang2022dtfd} with MaxMin feature selection (MaxMinS) and Aggregated feature selection (AFS) to assess the generalization of the proposed SC module to different MIL aggreagtors. 

\subsubsection{Experiment setup and Evaluation Metrics}
In this study, distinct experimental procedures were employed for different datasets.
Specifically, for the classical MIL datasets, we performed 10-fold cross-validation on each dataset with five repetitions per experiment, using classification accuracy as the primary metric. To validate the effectiveness of the proposed method, we incorporated the proposed SC module into the ABMIL framework using two distinct attention mechanisms, denoted by ABMIL \textbf{w/} SC and ABMIL-Gated \textbf{w/} SC. 

For CAMELYON16 dataset, we followed the protocols outlined in~\cite{zhang2022dtfd}.
Specifically, we randomly split the official training set into training and validation sets with a ratio of 90:10. We ran the experiments 5 times and reported the mean and standard deviation of all metrics.  For the TCGA-NSCLC dataset, we also followed~\cite{zhang2022dtfd} by performing a 4-fold cross-validation. This was done by splitting the entire dataset into training, validation, and testing sets with a ratio of 65:10:25. Similarly, the mean and standard deviation of all metrics were reported. 
The evaluation metrics used were the classification accuracy and the area under the receiver operating characteristic curve (AUC) scores.

\subsubsection{Implementation details}
The cross-entropy loss was adopted to train all the models in this work. The batch size was set to 1 for all the experiments. The models in classic MIL datasets were trained using the Adam optimizer for 40 epochs with initial learning of $1 \times 10^{-4}$ and $\ell_2$ weight decay of $5\times 10^{-3}$. The initial learning rate was adjusted through a cosine annealing scheduler. In the WSI classification tasks, we trained all four MIL aggregators for 200 epochs following the default training settings outlined in their official implementations~\cite{ilse2018attention, li2021dual, shao2021transmil, zhang2022dtfd}. It is worth noting that DSMIL is a special case, which used multi-scale feature embeddings obtained at both $\times 20$ and $\times 5$ magnification levels. Whereas, the remaining methods only used feature embeddings at $\times 20$ magnification level.
All experiments were implemented in \texttt{PyTorch} and performed on a Nvidia Tesla V100 GPU with 32G memory. 
\subsection{Results}
\subsubsection{WSI classification}
Integrating the proposed SC module consistently boosted the performance across four different types of MIL aggregators, using three feature extractors with different pretrained paradigms (see \Cref{tab:wsi}). The only exception was for TransMIL using CTransPath features, where we observed a slight drop of 0.4 \% in AUC after integrating the proposed SC module. 
This indicates that the performance gain of the proposed SC module is agnostic to different MIL aggregators and pretraining paradigms. On the CAMELYON16 dataset, plugging in the proposed SC module resulted in an average AUC gain of 2.63\%, 3.09\%, and 1.83\% across all MIL aggregators, using features extracted by ResNet-18, Swin-ViT, and CTransPath, respectively. Similarly, there was an average accuracy improvement of 2.29\%, 2.19\%, and 0.78\% when applying the SC module on these three feature sets. On the TCGA-NSCLC dataset, we observed an average increase of 1.11\%, 1.21\%, and 0.82\% in AUC using features extracted by ResNet-18, Swin-ViT, and CTransPath, respectively. An average improvement of 1.99\%, 1.84\%, and 1.83\% in accuracy was observed using the aforementioned features. 

In both the CAMEYLON16 and TCGA-NSCLC datasets, the improvement in the ABMIL-Gated aggregator was greater than the other MIL aggregators across three different feature sets (see~\Cref{tab:wsi}). This may be attributed to the fact that the ABMIL-Gated did not account for instance correlations, while the other MIL aggregators explicitly modeled instance correlations.
Consequently, we showed that integrating the proposed SC module into the ABMIL-Gated aggregator resulted in higher performance gain, as the SC module naturally captures instance correlations.
As consistent with findings in~\cite{lin2023interventional}, we observed that better feature embeddings generally led to better performance (CTransPath $>$ Swin-ViT $>$ ResNet-18). 
However, the performance gain of the integration of the SC module was in the opposite direction, with a higher performance gain for lower quality feature embeddings (\Cref{tab:wsi}). 
This suggests that enhancing high-quality feature embeddings is generally more challenging than a low-quality one for the proposed SC module. We would also like to point out that we did not observe a statistically significant difference in performance gains when integrating SC in networks using ResNet-18 and Swin-ViT feature sets, respectively. For ResNet-18, the increase of  AUC was 2.63\% on CAMELYON16, and 1.11\% on TCGA-NSCLC. For Swin-ViT, the AUC increase was 3.09\% on CAMELYON16 and 1.21\% on TCGA-NSCLC. This is because these two feature
extractors have similar performance and use the same pretraining paradigms on ImageNet.

\begin{figure*}[!t]
    \centering
    \includegraphics[width=0.95\textwidth]{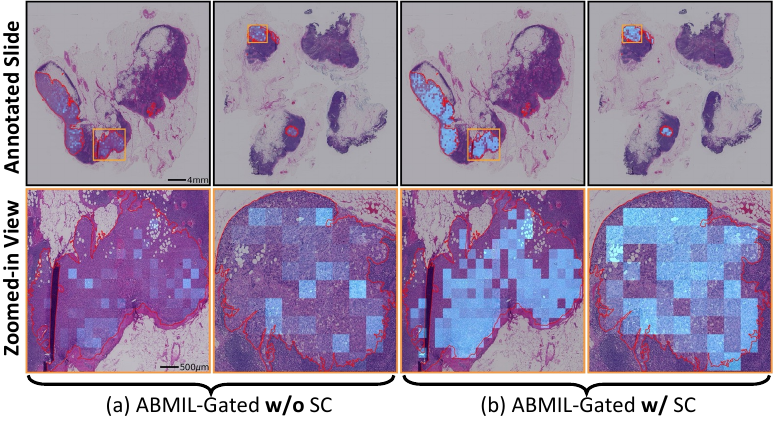}
    \caption{The tumor localization on the CAMELYON16 using ABMIL-Gated aggregator: (a) the attention map form ABMIL-Gated \textbf{w/o} SC, and (b) the attention map form ABMIL-Gated \textbf{w} SC. The \textcolor{red}{red} contours denote the ground-truth annotations of tumors. Each \textcolor{cyan}{blue} square represents the attention score for each WSI patch, where a brighter color signifies a higher attention score. }
    \label{fig:attn_wsi}
\end{figure*}
\begin{figure*}[!t]
    \centering
    \includegraphics[width=1.0\textwidth]{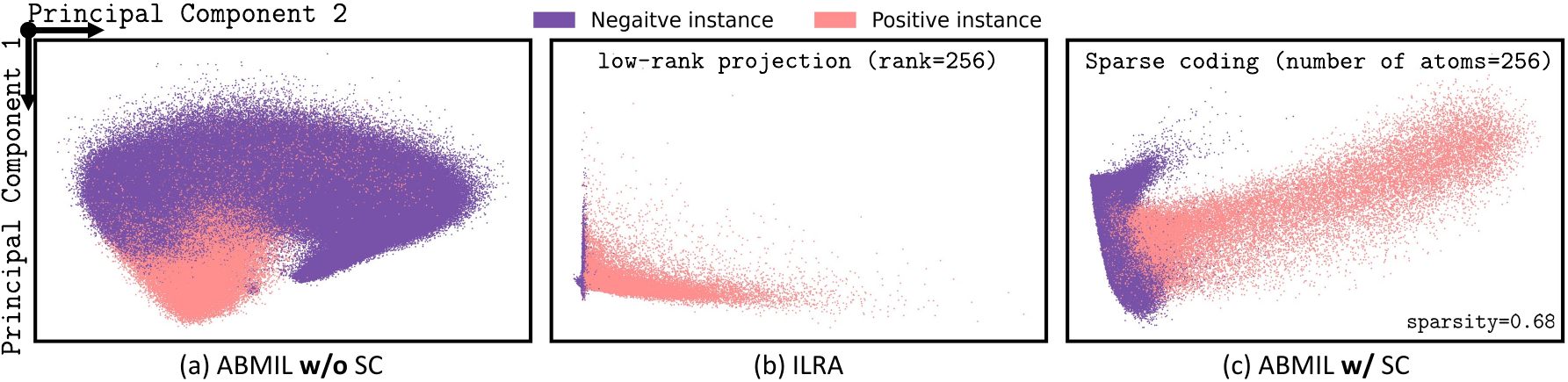}
    \caption{Visualization of the instance-level feature space using features extracted by a ResNet-18 on the CAMELYON16 testing set: (a) 512-dimensional features from after the first linear layer of a standard ABMIL; (b) 256-dimensional low-rank features from ILRA; (c) 256-dimensional sparse coefficients after performing the proposed SC module of an ABMIL.}
    \label{fig:pca}
\end{figure*}

\subsubsection{Classic MIL benchmarks}
Integrating the proposed SC module into ABMIL (ABMIL \textbf{w/} SC) and ABMIL-Gated (ABMIL-Gated \textbf{w/} SC) resulted in an average performance gain of 12.7\% and 13.6\% in classification accuracy, respectively, across five benchmark datasets. This improvement was determined to be statistically significant with $p<0.05$. In addition, ABMIL \textbf{w/} SC and ABMIL-Gated \textbf{w/} SC 
outperformed the previous state-of-the-art methods across all five MIL benchmark datasets regarding classification accuracy (see \Cref{tab:classic}). The ABMIL-Gated \textbf{w/} SC achieved the best performance by improving the previous state-of-the-art accuracy by an average of 4.95\%, with 3.97\% on MUSK1, 3.23\% on MUSK2, 8.50\% on FOX, 5.69\% on TIGER, and 3.35\% on ELEPHANT. Moreover, the ABMIL-Gated \textbf{w/} SC exhibited the highest stability with an average standard deviation of 0.0054 in classification accuracy.   

\subsection{Ablation on Model Design Variants}
\subsubsection{Analysis on hyperparameters}
We conducted ablation studies to investigate the impact of two key hyperparameters in the proposed SC module (i.e., number of atoms $m$ in the dictionary and number of unrolled layers $L$) on performance. The ablations were conducted on the CAMELYON16 dataset using features extracted by a ResNet-18 and ABMIL-Gated aggregator; unless specified otherwise.

We first examined the impact of the number of atoms by maintaining $L=5$. We observed that increasing the number of atoms resulted in a gradual improvement in performance as well as an increase in parameters and computation (\Cref{tab:atoms}). We noticed that increasing the number of atoms from 256 to 512 only resulted in a minor performance gain of 0.10\% in AUC, while the computational cost increased approximately ten times. To investigate the effect of the number of unrolled layers on the performance, we fixed $m=256$. Overall, increasing the number of unrolled layers progressively led to an improvement in AUC, but at the expense of increased computational cost (\Cref{tab:layers}). A drop was observed when increasing $L$ from 5 to 7 and from 7 to 9, which may be attributed to minor fluctuation in the convergence path of LISTA. Therefore, we reported the results using $L=5$ and $m=256$ to balance performance and computational cost.

\subsubsection{Sparse coding vs low-rank projection}
We compared the proposed sparse coding (number of atoms $m$ = 256) with the low-rank projection used in ILRA~\cite{xiang2023exploring} (rank = 256) on the CAMELYON16 dataset. As shown in~\Cref{fig:sc_ilra}, the proposed SC module consistently showed superior performance compared to the low-rank projection (i.e., ILRA) using features extracted by either a ResNet-18 or a Swin-ViT. It is worth noting that the ILRA is tailored to the transformer-based MIL aggregator. While the proposed SC module can be plugged into any existing MIL aggregator. We observed that the ABMIL-Gated w/ SC surpassed the transformer-based ILRA by 1.54\%. Similarly, the TransMIL w/ SC outperformed ILRA by 2.60\%.
We hypothesized that the superior performance of SC may be attributed to the over-complete dictionary. The over-complete dictionary offers a more compact and robust way to capture the similarity and variability among instances than dense representations provided by a low-rank projection.

\subsection{Interpretation}
\subsubsection{Localization performance} We quantified the performance of localization in terms of Free-Response Receiver Operating Characteristic (FROC), which is computed as the average sensitivity of detection at 6 predefined numbers of false positives rates per slide: 1/4, 1/2, 1, 2, 4 and 8. As shown in \Cref{tab:localization}, integrating the proposed SC module to ABMIL-Gated improved the FROC by 53.7\% on the CAMELYON-16 test set using ImageNet features. 
As shown in \Cref{fig:attn_wsi}, vanilla ABMIL-Gated exhibited poor tumor localization, missing most of the tumor patches. Whereas, the integration of the SC module enhanced the localization performance of the ABMIL-Gated, aligning well with the ground-truth annotation. The findings evidence that the global dictionary of instance embeddings in the proposed SC module can effectively capture cross-instance similarities, leading to enhanced localization performance. 

\begin{table}[!t]
\centering
\caption{Comparison of localization performance in terms of average FROC ($\pm$ standard deviation) on the CAMELYON-16 test set using features extracted by ResNet-18 pretrained on ImageNet. Integrating the proposed SC module led to a better localization performance.}
\label{tab:localization}
\resizebox{0.45\textwidth}{!}{
\begin{tabular}{c|cc}
\toprule %
 Method  &  Probability Map From & FROC \\ 
 \midrule
 ABMIL-Gated \textbf{w/o} SC & attention score & 0.246 $\pm$ 0.022 \\
 ABMIL-Gated \textbf{w/} SC & attention score & \textbf{0.378} $\pm$ 0.025 \\
 \bottomrule
\end{tabular}%
}
\end{table}

\subsubsection{Learned instance-level representation}
We visualized the instance-level feature space learned in standard ABMIL, ILRA, and ABMIL w/ SC for both positive and negative instances in the CAMELYON16 test set using ImageNet features. For this purpose, the high-dimensional feature space was reduced to 2D space using principal component analysis (PCA). First, we observed that the principal component representations of negative and positive instances learned in ILRA and ABMIL w/ SC were easier to discriminate compared to those learned in the standard ABMIL (see~\Cref{fig:pca}). This may contribute to their superior performance compared to standard ABMIL. Second, we observed that the representations of both positive and negative instances learned in ILRA were concentrated along a slender line. Whereas, those in ABMIL w/ SC spanned a wider space (see~\Cref{fig:pca}(b) and (c)). 
This is because the over-complete dictionary in the proposed SC module better captured the variability among positive and negative instances compared to the low-rank projection used in ILRA.

\section{Conclusion}
In this paper, we proposed a novel MIL framework, termed SC-MIL, by leveraging unrolled sparse dictionary learning. 
The proposed method simultaneously enhances the instance feature embedding and models cross-instance similarities without significantly increasing the computational complexity. Importantly,  experimental results from multiple benchmarks across various tasks showed that the performance of state-of-the-art MIL methods could be further boosted by incorporating the proposed SC module in a plug-and-play manner. The proposed method exhibits great potential to be used in real-world applications to aid in drug effect prediction and the diagnosis and pathological analysis of cancers using histology. The proposed method is particularly effective in real scenarios where the self-supervised pre-training for the enhancement of feature embedding is infeasible due to limited data size.

\subsubsection*{Limitation} Although the unrolled sparse dictionary learning can automatically handle expensive hyperparameter tuning in traditional iterative solutions, a limitation of the proposed method is that it necessitates minor hyperparameter tuning, e.g., the number of atoms and learning rate. In addition, plugging the SC module into existing MIL frameworks may result in a slight slowdown in convergence. We will explore these limitations in future work. 

\appendices

\section*{Acknowledgment}
Computations were performed using the facilities of the Washington University Research Computing and Informatics Facility (RCIF). The RCIF has received funding from NIH S10 program grants: 1S10OD025200-01A1 and 1S10OD030477-01.

\bibliography{refs.bib}
\bibliographystyle{IEEEtran}

\end{document}